%% file: main.tex
\documentclass[conference]{IEEEtran}
\IEEEoverridecommandlockouts   

\input{preamble}

\begin{document}
\makeatletter
\def\@IEEEauthorblockconfadjspace{-18pt}%
\makeatother

\title{Reasoning Without Inference Cost: Latent Semantic\\
Scaffolding for Robot VLA Policies}

\author{Andrew Ting Yan Li, Zhuo Li, Zhelin Yang, Zhipeng Dong, Quentin Rouxel,
and Fei Chen,~\textit{Senior Member, IEEE}%
\thanks{$^{*}$This work is supported in part by the Research Grants Council of the
Government of the Hong Kong SAR via the Grants 24209021, 14213324, C7100-22GF,
and in part by the InnoHK of the Government of the Hong Kong SAR via the Hong
Kong Centre for Logistics Robotics. \textit{(Corresponding author: Fei Chen.)}}%
\thanks{All authors are with the Collaborative and Versatile Robots (CLOVER)
Laboratory, T-Stone Robotics Institute, The Chinese University of Hong Kong,
Hong Kong (e-mail: ali@mae.cuhk.edu.hk; zli@mae.cuhk.edu.hk;
zlyang@mae.cuhk.edu.hk;
zhipengdong@cuhk.edu.hk; 
quentinrouxel@cuhk.edu.hk; f.chen@ieee.org).}%
}

%
\IEEEaftertitletext{%
\vspace{-1.2em}
\setlength{\abovecaptionskip}{2pt}%
\noindent
\begin{minipage}[t]{0.48\textwidth}
  \vspace{0pt}
  \small\bfseries
  \textbf{Abstract}---Vision-language-action (VLA) models are trained by
  imitation and capture \emph{what} action to take but not \emph{why}; adding
  causal reasoning improves manipulation, but current methods pay for it at
  inference time---generating reasoning tokens or rolling out predicted future
  states at every step, a cost that compounds over long horizons. We ask whether
  this benefit can instead be captured during training and discarded before
  deployment. We introduce \textbf{Latent Semantic Scaffolding (\lss)}, an
  auxiliary loss applied during human-demonstration pretraining that aligns a
  VLA's action-token representations to text embeddings of physical-reasoning
  rationales through a small projection head. The head is dropped at inference,
  leaving the unmodified base policy with \emph{zero} added cost. Our central
  finding concerns \emph{alignment granularity}: aligning each action token to the
  rationale of its own manipulation phase (\dense) rather than to a single pooled
  episode-level embedding (\pooled) yields representations that transfer markedly
  better to held-out tasks. \dense{} attains both the best in-distribution success
  and the best transfer to tasks unseen during alignment, whereas pooled alignment
  over-specializes to the training task. A representational probe shows \dense{}
  induces roughly twice the per-phase separability in the backbone, supporting that
  phase-local alignment is the operative mechanism.
\end{minipage}\hfill
\begin{minipage}[t]{0.48\textwidth}
  \vspace{0pt}
  \centering
  \includegraphics[width=1.05\linewidth]{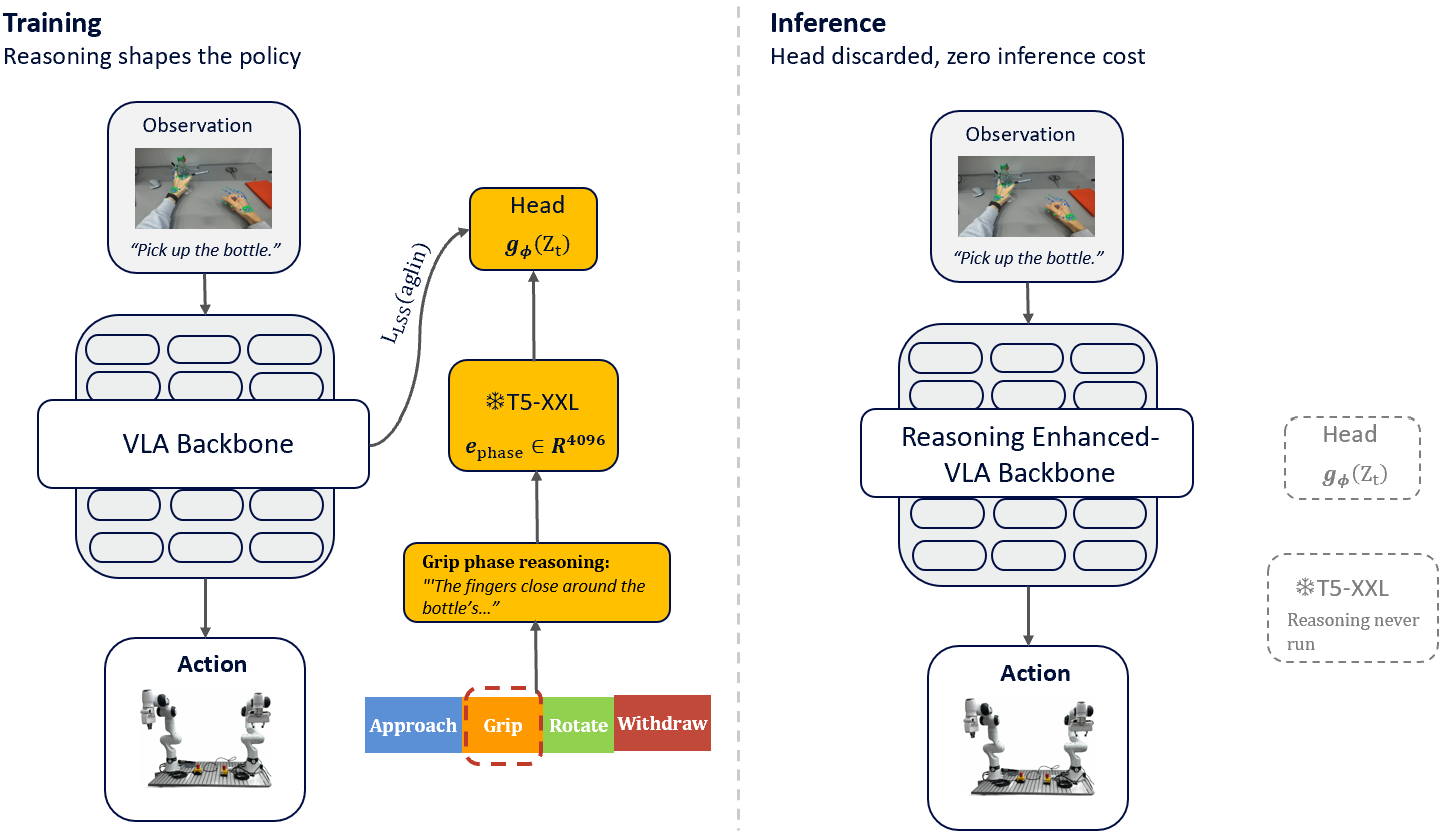}
  \captionof{figure}{\textbf{\lss{} injects reasoning at training time and discards
  it at inference.} \emph{Left (training):} the backbone is trained on the
  flow-matching objective while an auxiliary projection head $g_\phi$ aligns each
  action token to the T5 embedding of its manipulation-phase reasoning---an
  auxiliary branch that shapes the backbone but does not lie on the
  action-production path. \emph{Right (inference):} the head and reasoning encoder
  are removed; the deployed policy is the unmodified base VLA with reasoning
  ``baked in,'' at zero added cost.}
  \label{fig:teaser}
\end{minipage}%
}

\maketitle
\thispagestyle{empty}

\input{sections/intro}
\input{sections/related}
\input{sections/method}
\input{sections/experiments}
\input{sections/limitations}
\input{sections/conclusion}

\bibliographystyle{IEEEtran}
\bibliography{references}

\end{document}

%% file: preamble.tex
\usepackage{cite}
\usepackage{amsmath,amssymb,amsfonts}
\usepackage{algorithmic}
\usepackage{algorithm}     
\usepackage{graphicx}
\usepackage{textcomp}
\usepackage{xcolor}
\usepackage{booktabs}     
\usepackage{multirow}
\usepackage{array}
\usepackage{subcaption}   
\usepackage{url}
\usepackage[section]{placeins}  
\usepackage{cuted}        

\graphicspath{{figures/}}   

\usepackage[colorinlistoftodos,prependcaption,textsize=small]{todonotes}

\newcommand{\figref}[1]{Fig.~\ref{#1}}
\newcommand{\tabref}[1]{Table~\ref{#1}}
\newcommand{\secref}[1]{Sec.~\ref{#1}}

\newcommand{\lss}{LSS}
\newcommand{\dense}{Dense \lss}
\newcommand{\pooled}{Pooled \lss}
\newcommand{\head}{LSAHead}
\newcommand{\hrdt}{H-RDT}

\newcommand{\adjustbottle}{\texttt{adjust\_bottle}}
\newcommand{\shakebottle}{\texttt{shake\_bottle}}
\newcommand{\movecanpot}{\texttt{move\_can\_pot}}

\newcommand{\Zact}{Z_{\text{act}}}
\newcommand{\Lflow}{L_{\text{flow}}}
\newcommand{\Llss}{L_{\text{LSS}}}
\newcommand{\Ldense}{L_{\text{dense}}}
\newcommand{\etext}{e_{\text{text}}}
\newcommand{\R}{\mathbb{R}}

\def\BibTeX{{\rm B\kern-.05em{\sc i\kern-.025em b}\kern-.08em
    T\kern-.1667em\lower.7ex\hbox{E}\kern-.125emX}}

%% file: sections/intro.tex
\vspace*{0.0005\baselineskip}

\section{Introduction}
Vision-language-action models (VLAs) map perception to action by imitation,
learning generalizable sensorimotor control from large-scale robot and
non-robot demonstrations\cite{rt2,openvla,pi0,droid}. To scale beyond the cost of
teleoperated robot data, a growing line of work pretrains VLAs on egocentric
\emph{human} demonstrations before finetuning on a small amount of robot
data\cite{hrdt,egovla,beingh0}. Yet a policy trained purely to
reproduce demonstrated actions captures the \emph{what} of a motion without its
causal \emph{why}. Recent work shows that injecting reasoning---an explicit
account of the task's causal and physical structure---improves manipulation,
particularly on tasks that require multiple steps or generalization beyond the
training distribution\cite{cotvla,ecot}.

This benefit, however, is typically realized at \emph{inference} time. Embodied
chain-of-thought policies generate textual reasoning traces autoregressively
before each action\cite{ecot}, while visual chain-of-thought policies predict
future image frames as subgoals and then act toward them\cite{cotvla}. World
action models go further still, jointly predicting future states and actions in a
unified model so that action generation is coupled to an explicit forecast of the
next state\cite{worldvla}. Both families re-incur their reasoning or prediction computation at every
control step, and the cost is large: a recent robustness study reports that world
action models run $4.8\times$ to $83\times$ slower per action chunk than a
comparable flow-based VLA\cite{worldmodel}, and dedicated efficiency work exists
purely to reduce the overhead of autoregressive reasoning
traces\cite{fastecot}. The cost also compounds with task
length---a
long-horizon task re-pays the reasoning tax at every sub-step, precisely where
multi-step planning matters most.

Standard VLA supervision is also coarse: a whole trajectory is paired with a
single instruction phrase, with no link between language sub-parts and action
sub-segments\cite{openx}. Fine-grained datasets exist, but typically treat
sub-tasks as prediction or segmentation labels rather than causal
rationale\cite{roboactclip}.

This raises the question we study: \emph{must} reasoning's benefit be paid at
inference at all, or can it be captured at training time and discarded? We
hypothesize the latter---if reasoning helps by shaping a policy's internal
representations rather than by being generated step by step, its benefit can be
baked into the weights, and the runtime cost avoided.

We realize this with \lss{} and, crucially, study \emph{how} the alignment
should be structured (Fig.~\ref{fig:teaser}). Our contributions are:
\begin{itemize}
  \item \textbf{\lss}, a training-time auxiliary loss that aligns a VLA's
        action-token representations to embeddings of physical-reasoning text.
        The alignment head is discarded at inference, adding zero cost over the
        base policy.
  \item \textbf{An alignment-granularity result}: per-token, phase-local
        (\emph{dense}) alignment transfers to held-out tasks better than
        episode-pooled alignment. A controlled finetuning ladder isolates this
        from the human-data and task-matching confounds.
  \item \textbf{A representational probe} showing dense alignment induces
        ${\sim}2\times$ stronger per-phase separation in the backbone---supporting
        evidence that phase-local shaping drives the transfer gain.
\end{itemize}

%% file: sections/related.tex
\vspace*{0.5\baselineskip}

\section{Related Work}

\subsection{Test-time reasoning and prediction}
A growing line of work improves manipulation by having the policy reason
\emph{before} it acts, generating an intermediate artifact at inference time.
Embodied chain-of-thought (ECoT) trains a VLA to emit step-by-step textual
reasoning---plans, sub-tasks, and visually grounded features such as object
bounding boxes and end-effector positions---before predicting each
action\cite{ecot}. Visual chain-of-thought instead predicts future image frames
as visual subgoals and then produces a short action sequence toward
them\cite{cotvla}. These differ in the \emph{modality} of what they generate
(text vs.\ predicted pixels) but share a defining trait: the reasoning
computation recurs at every inference step. Because the traces are produced
autoregressively, this becomes the dominant runtime cost, motivating dedicated
efficiency work that reuses or parallelizes the reasoning to recover
speed\cite{fastecot}. The overhead compounds over long horizons. Our work moves
this computation entirely to training time, so the deployed policy carries none
of it.

\subsection{Representation alignment as a training signal}
A separate line uses representation alignment as an auxiliary training objective
rather than a generation target. In image generation, REPA regularizes a
diffusion transformer by aligning its noisy hidden states to features from a
pretrained \emph{visual} encoder, substantially accelerating
convergence\cite{repa}. This is a close methodological relative of our loss---a
cosine alignment through a lightweight projection head---but in a different domain
(generation, not control) and with a different teacher (vision features, not
reasoning). We differ on both the \emph{teacher} and the \emph{granularity}: we
align action-token representations to embeddings of causal reasoning text, and we
show that aligning each token to its own manipulation-phase rationale, rather
than to a single pooled target, is what produces transferable representations.
Fine-grained sub-task datasets exist but treat sub-tasks as prediction or
segmentation labels\cite{roboactclip}; we
instead align to causal rationale and, unlike all of the above, discard the
alignment apparatus entirely at inference.

\subsection{Human-demonstration pretraining for VLAs}
Egocentric human manipulation data offers a scalable source of behavioral priors
for robot policies. H-RDT pretrains a 2B-parameter diffusion-transformer policy
on large-scale human manipulation data with paired 3D hand-pose annotations, then
cross-embodiment finetunes on robot data\cite{hrdt}; EgoVLA and Being-H0 likewise
learn from human video before light robot finetuning\cite{egovla,beingh0}. We
adopt H-RDT as our backbone. Its original recipe has two stages---human pretrain
followed by robot finetune---and we insert an \lss-supervised human-pretraining
stage between them. That inserted stage, not the human-demonstration paradigm
itself, is our contribution.

\lss{} thus occupies an intersection the above do not fill: \emph{alignment}
rather than generation (unlike CoT/world-model methods), a \emph{reasoning-text}
teacher rather than a vision teacher (unlike REPA-style alignment), inside
\emph{human-demonstration pretraining}---with phase-local structure carrying
causal rationale, at zero added inference cost.

%% file: sections/method.tex
\vspace*{0.5\baselineskip}
\section{Method}

\begin{figure*}[!tbp]
  \centering
  \includegraphics[width=0.8\linewidth]{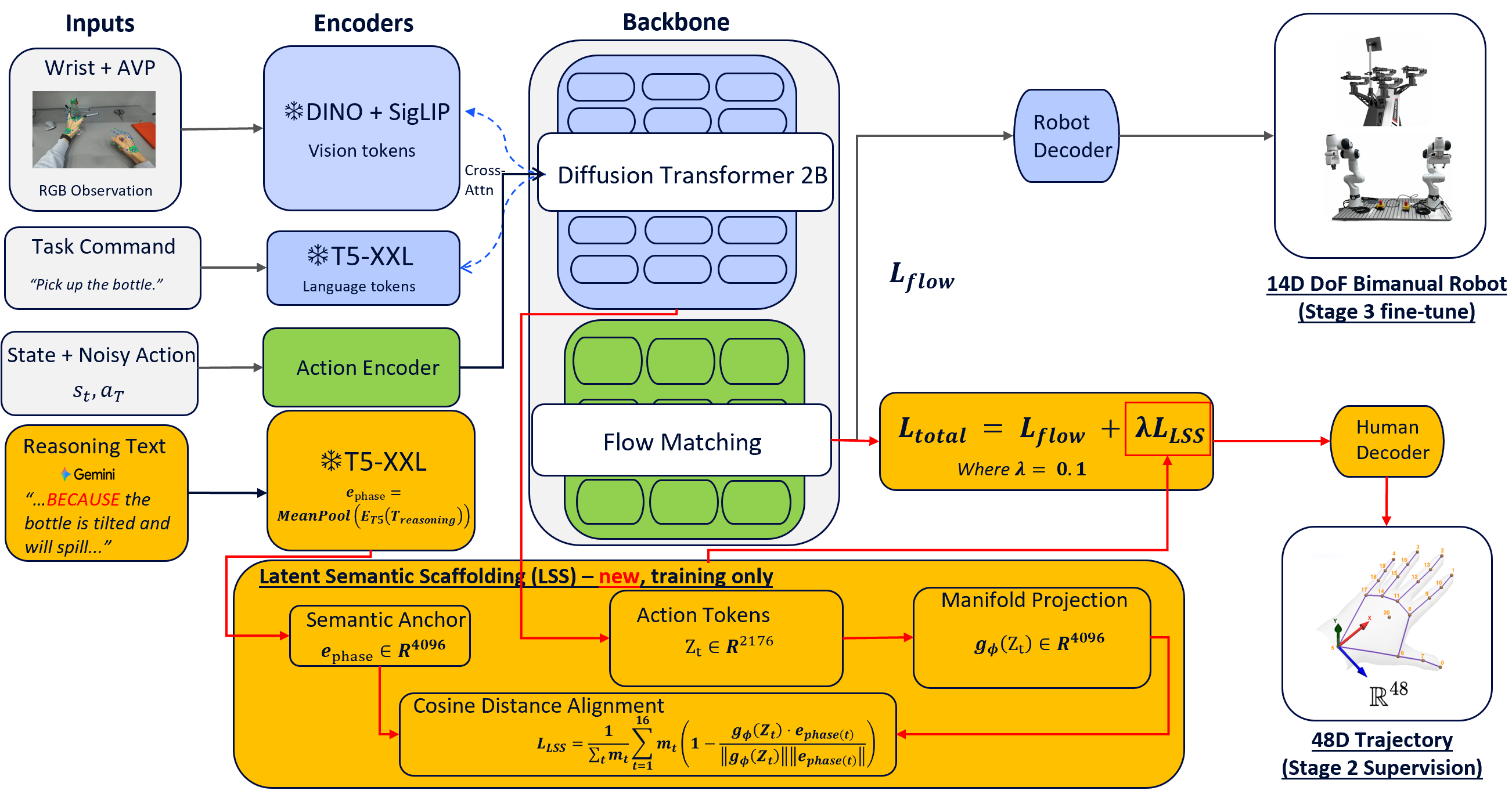}
  \caption{\textbf{\dense{} architecture (Stage~2, training only).} A frozen
  DINO+SigLIP encoder and a frozen T5-XXL command encoder condition the 2B
  diffusion-transformer backbone, which produces $K{=}16$ action tokens
  $Z_t\in\R^{2176}$ supervised by the flow-matching loss $\Lflow$. In parallel,
  the \head{} projection $g_\phi$ maps each action token into T5 space and a
  cosine-distance loss aligns it to the reasoning embedding of its own
  manipulation phase, $e_{\text{phase}(t)}$ (straddling tokens masked by $m_t$).
  The combined objective is $L_{\text{total}}=\Lflow+\lambda\Llss$. \pooled{} uses
  the identical architecture but replaces the per-phase targets with a single
  mean-pooled episode embedding (\eqref{eq:pooled}). The entire \lss{} apparatus
  (orange) is discarded at inference, leaving the unmodified base policy.}
  \label{fig:arch}
\end{figure*}

\subsection{Preliminaries}
We build on H-RDT\cite{hrdt}: DINO+SigLIP vision encoders, a frozen T5-XXL encoder
for the task command, a 2B-parameter diffusion-transformer backbone $f_\theta$,
and a flow-matching action head. Conditioned on an image observation and the task
command, the backbone produces $K{=}16$ action tokens
$\Zact = \{z_1,\dots,z_K\} \in \R^{K\times 2176}$, decoded by the action head
into a trajectory; the primary objective is the flow-matching loss $\Lflow$.
H-RDT's original recipe has two stages: human-data pretraining and robot
finetuning. We insert a second pretraining stage in which \lss{} is active,
giving a three-stage pipeline: (i)~human pretrain, (ii)~\lss{} human pretrain,
(iii)~robot finetune. Stage~2 supervises H-RDT's unified 48-D human action
representation; Stage~3 finetunes to the 14-D robot.

\subsection{Reasoning as an alignment target, not an input}
Each human trajectory carries a physical-reasoning description $T_{\text{reason}}$
generated offline by a VLM (\secref{sec:annot}) and encoded by a \emph{frozen}
T5-XXL. \textbf{The reasoning text is used only as the alignment target---it never
conditions the policy.} Language conditioning comes solely from the task command;
reasoning influences the policy purely through the \lss{} loss, and the
reasoning-T5 embeddings are precomputed and cached. Consequently the reasoning
encoder is never instantiated at inference.

\subsection{Pooled \lss{}}
\lss{} adds a non-linear projection head, the \head{}
$g_\phi:\R^{2176}\!\to\!\R^{4096}$, mapping action-token hidden states into the
T5 embedding space. In the pooled variant, the action tokens are mean-pooled to
$\bar z = \tfrac{1}{K}\sum_i z_i$ and aligned to a single mean-pooled reasoning
embedding $\etext=\mathrm{MeanPool}(E_{T5}(T_{\text{reason}}))$ by a cosine
distance:
\begin{equation}
\Llss = 1 - \frac{g_\phi(\bar z)\cdot \etext}
{\lVert g_\phi(\bar z)\rVert\,\lVert \etext\rVert},
\qquad
L_{\text{total}} = \Lflow + \lambda\,\Llss,
\label{eq:pooled}
\end{equation}
with $\lambda{=}0.1$. The \head{} is a two-layer MLP
($2176\!\to\!2176$, GELU, $2176\!\to\!4096$). The alignment loss acts as a
secondary pressure (small $\lambda$) that reshapes action representations toward
the reasoning structure without overriding action prediction. The full
architecture is shown in \figref{fig:arch} (the figure depicts the dense variant;
pooled differs only in the alignment target).

\subsection{Dense \lss{} (phase-local alignment)}
\label{sec:dense}
\begin{figure*}[!t]
  \centering
  \includegraphics[width=0.8\linewidth]{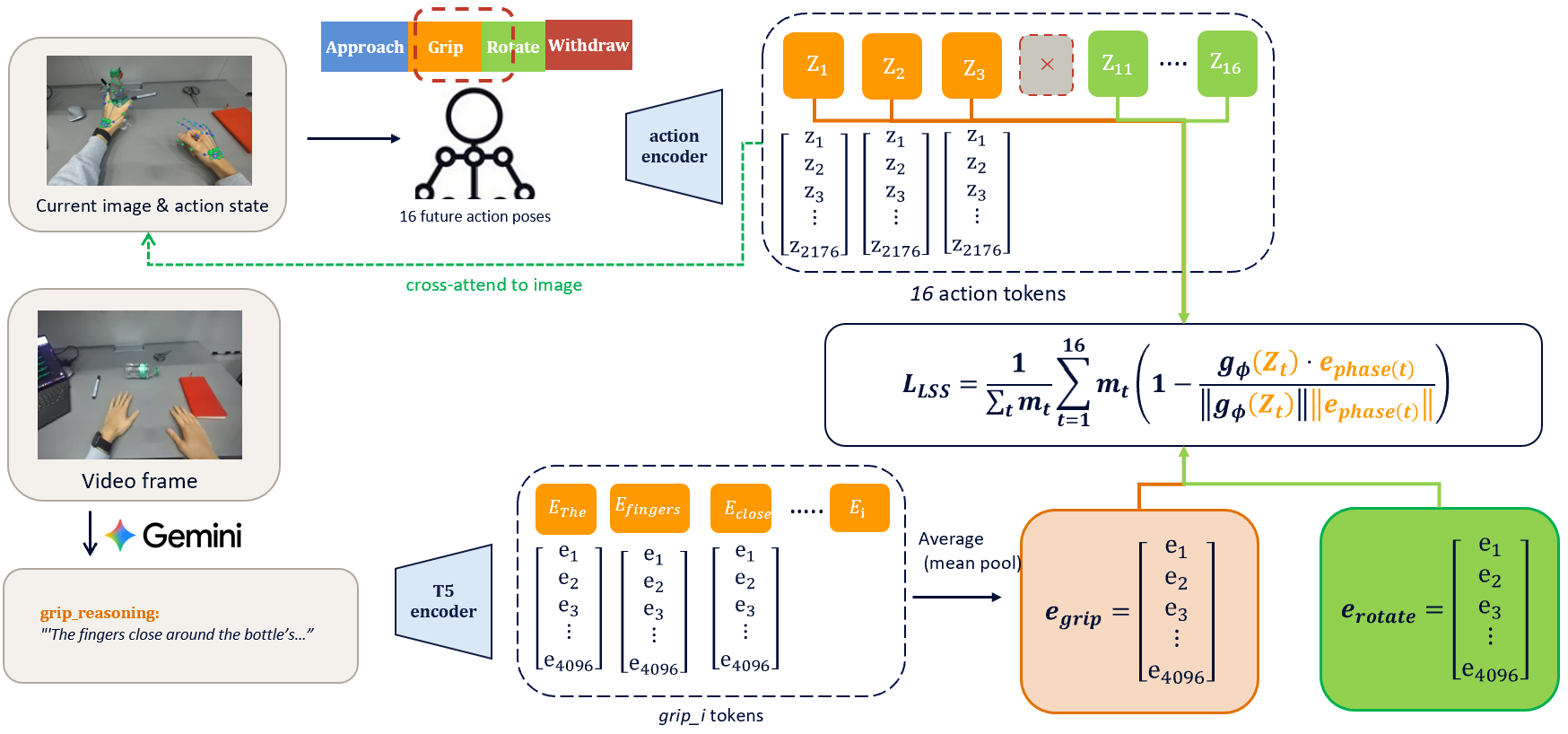}
  \caption{\textbf{\dense{} token-to-phase alignment.} Each of the 16 action
  tokens $Z_t$ is assigned to the manipulation phase spanned by its frames
  (tokens straddling a boundary, e.g. the greyed slot, are masked by $m_t$);
  color indicates the assigned phase (orange = grip, green = rotate; two
  representative phases shown). The per-phase reasoning text is encoded by a frozen
  T5 and mean-pooled into a target $e_{\text{phase}(t)}$. The head $g_\phi$
  projects each token into T5 space and $\Ldense$ aligns it to the target of its
  own phase.}
  \label{fig:dense}
\end{figure*}

Mean-pooling collapses both the $K$ action tokens and the multi-phase rationale
into single vectors, discarding temporal order and the per-phase causal content.
\dense{} preserves this structure. The rationale is segmented into $P$ ordered
phases from a fixed vocabulary (\textit{approach}, \textit{grip}, \textit{rotate},
\textit{withdraw}), each carrying its own one-sentence causal rationale and a
frame range, and each T5-encoded and masked-mean-pooled into a per-phase target
$e_p$. Each action token spans a fixed number of frames
(\texttt{upsample\_rate}${=}3$) and is assigned to the phase those frames fall in;
the ${\sim}5\%$ of tokens whose frames straddle a phase boundary are masked out of
the loss. The \head{} is then applied \emph{per token} and each token aligned to
the target of its own phase:
\begin{equation}
\Ldense = \frac{1}{\lvert\mathcal{M}\rvert}\sum_{i\in\mathcal{M}}
\left[\,1 - \frac{g_\phi(z_i)\cdot e_{p(i)}}
{\lVert g_\phi(z_i)\rVert\,\lVert e_{p(i)}\rVert}\,\right],
\label{eq:dense}
\end{equation}
where $p(i)$ is token $i$'s phase and $\mathcal{M}$ is the clean (non-straddling)
token set. The cosine objective is identical to \eqref{eq:pooled}; only the
target changes, from one pooled embedding to a phase-local one
(Fig.~\ref{fig:dense}). The pooled and
dense paths are otherwise interchangeable. \emph{Caveat:} \dense{} depends on
reliable phase boundaries---noisy segmentation can misalign tokens and underperform
pooling, which is why pooled-vs-dense is an empirical question rather than a
foregone win.

\subsection{Training-only design; zero inference overhead}
The entire \lss{} apparatus---the \head{}, the reasoning-T5, and the alignment
loss (pooled or dense)---is used only in training. The \head{} is never saved with
the model and is never instantiated at inference; the reasoning encoder is never
run. The deployed computation graph is exactly the unmodified base VLA, conditioned
on a plain task instruction, so \lss{} adds \textbf{zero cost} over the base
policy. Reasoning's influence persists only in the trained backbone weights, not in
any runtime module. Algorithm~\ref{alg:lss} summarizes the procedure.

\begin{algorithm}[!t]
\caption{Stage-2 training with Dense \lss{}}
\label{alg:lss}
\begin{algorithmic}[1]
\STATE \textbf{Input:} backbone $f_\theta$, head $g_\phi$, frozen T5 $E_{T5}$,
       phase-annotated human data $\mathcal{D}$, weight $\lambda$
\STATE \textbf{Precompute:} per-phase targets
       $e_p \gets \mathrm{MeanPool}(E_{T5}(T_p))$ for all phases, cache
\FOR{each minibatch $(x, a, \text{phases}) \sim \mathcal{D}$}
    \STATE $Z \gets f_\theta(x)$ \quad // 16 action tokens
    \STATE $L_{\text{flow}} \gets$ flow-matching loss on $Z, a$
    \STATE assign each token $z_i$ to its phase $p(i)$; mask straddlers $\to \mathcal{M}$
    \STATE $\Ldense \gets \frac{1}{|\mathcal{M}|}\sum_{i\in\mathcal{M}}
           \big[1 - \cos(g_\phi(z_i), e_{p(i)})\big]$
    \STATE update $\theta,\phi$ by $\nabla(L_{\text{flow}} + \lambda\,\Ldense)$
\ENDFOR
\STATE \textbf{At inference:} discard $g_\phi$ and $E_{T5}$; deploy $f_\theta$ alone
\end{algorithmic}
\end{algorithm}

\subsection{Reasoning annotation pipeline}
\label{sec:annot}
Per-trajectory rationales are generated offline by Gemini-3-flash, prompted to
explain the \emph{causal} structure of the manipulation---the object-state trigger
and the physical response---rather than to narrate the motion. For \dense{}, the
model additionally returns the phase segmentation (phase labels with time-fraction
ranges), later converted to action-frame ranges. Phase boundaries are validated
primarily by VLM-vs-human video comparison; a kinematic cross-check is recorded as
a diagnostic only. We apply no formal quality control on the generated text, which
we flag as a limitation.

\FloatBarrier
\vspace*{0.33\baselineskip}

%% file: sections/experiments.tex
\section{Experiments}
\subsection{Setup}
We evaluate on RoboTwin~2.0~\cite{robotwin2} with the aloha-agilex 14-DOF configuration
(Fig.~\ref{fig:robotwin}). Following the benchmark's standard protocol, every
Stage-3 finetune uses an identical budget of 50 HDF5 episodes and 22{,}000 steps,
so that runs differ only in the Stage-2 backbone they start from. Success rates
are reported over 100 rollouts.

\subsection{Data collection}
\begin{figure}[t]
  \centering
  \begin{subfigure}{\linewidth}
    \centering
    \includegraphics[width=0.95\linewidth]{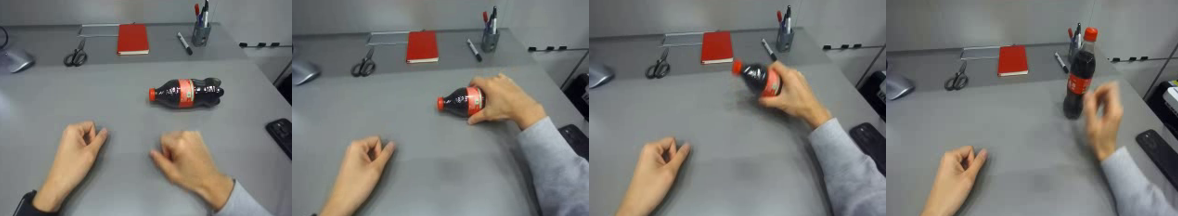}
    \caption{One \adjustbottle{} human demonstration (right hand), shown as its
    four manipulation phases: approach, grip, rotate, withdraw. Teleoperated via
    Apple Vision Pro; these phases define the dense alignment targets.}
    \label{fig:dataset}
  \end{subfigure}
  \\[1ex]
  \begin{subfigure}{\linewidth}
    \centering
    \includegraphics[width=0.95\linewidth]{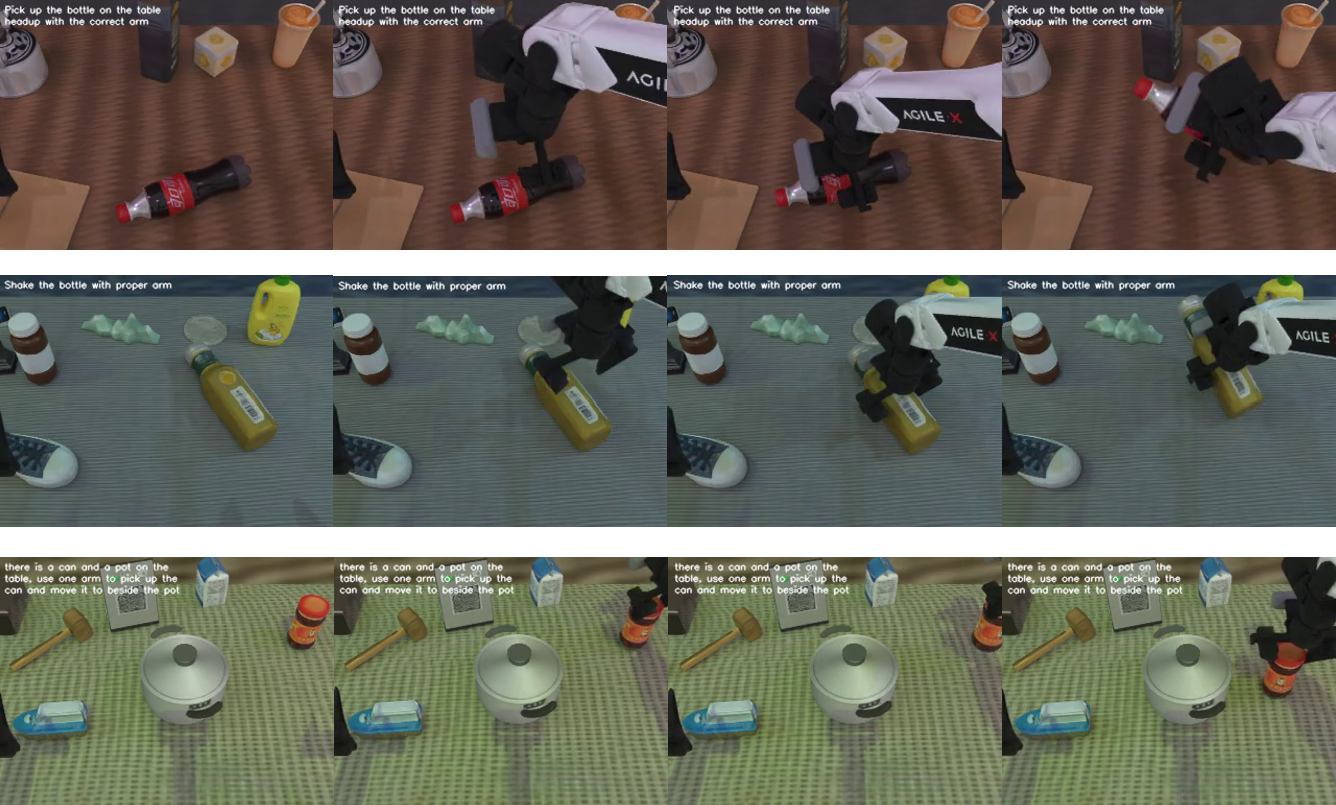}
    \caption{RoboTwin~2.0 evaluation tasks (aloha-agilex, 14-DOF), shown as four
    frames per task. The three tasks form a similarity gradient relative to the
    \adjustbottle{} human pretraining data: \adjustbottle{} (top, in-distribution),
    \shakebottle{} (middle, a related bottle-manipulation task), and \movecanpot{}
    (bottom, a structurally dissimilar task), probing near- and far-transfer from
    the same Stage-2 backbone.}
    \label{fig:robotwin}
  \end{subfigure}
  \caption{Human demonstration data (top) and simulation evaluation
  environment (bottom).}
\end{figure}

Human demonstrations are collected on a teleoperation rig that produces data in
the EgoDex format~\cite{egodex}. An Apple Vision Pro records the operator's 48-D
hand and wrist
motion, while a separate head-mounted ZED stereo camera captures the front-view
RGB stream. The ZED is required because Apple does not open-source the Vision
Pro's on-device recording module, so pose and video are captured by distinct
devices and synchronized in our data pipeline, which builds on the
human-policy framework of~\cite{humanpolicy}. Using this rig we collected
500 \adjustbottle{} episodes (Fig.~\ref{fig:dataset}) in a balanced $2\times2$
design---125 episodes for each combination of hand (left, right) and bottle
instance (Coke, Sprite)---yielding 250 left-hand and 250 right-hand
demonstrations. Each episode is processed into the unified 48-D human action
representation used for Stage-2 pretraining; reasoning text is Gemini-generated
causal rationale.

\subsection{Main ablation ladder}
\tabref{tab:ablation} isolates the \lss{} mechanism on the in-distribution task.
Two comparisons are load-bearing. First, aligning to a trivial instruction-text
target collapses performance (58\%) far below aligning to genuine reasoning
(\pooled, 85\%): the \emph{content} of the alignment target, not the mere presence
of an auxiliary loss, drives the gain. Second, supplying the same reasoning text as
a language input (Run~C, 71\%) or via a token-prediction loss (Run~D, 74\%)
underperforms embedding alignment---reasoning helps as an alignment signal, not as
generated or conditioned text. Run~C even falls below the AVP-only baseline
(Run~B, 79\%), consistent with reasoning text acting as distracting input when the
model is not given an objective that uses it. \dense{} attains the best result
(90\%), establishing alignment granularity as a further lever beyond reasoning
content. Run~F (frozen backbone) confirms the priors live in the backbone weights,
not the discarded head.

\begin{table}[h]
\centering
\caption{Ablation ladder on \adjustbottle{} (RoboTwin~2.0,
\texttt{demo\_randomized}, 100 rollouts). The 58\%\,$\to$\,85\% gap (trivial
target vs.\ Run~E) shows the alignment \emph{target} matters, not merely the
presence of an auxiliary loss.}
\label{tab:ablation}
\small
\setlength{\tabcolsep}{4pt}
\begin{tabular}{@{}clccc@{}}
\toprule
Run & Mechanism & AVP & Reas. & Succ.\,(\%) \\
\midrule
A  & EgoDex pretrain + finetune (baseline) & --         & --         & 70 \\
B  & AVP kinematics, no reasoning          & \checkmark & --         & 79 \\
C  & Reasoning as language input           & \checkmark & \checkmark & 71 \\
D  & Reasoning via token-prediction loss   & \checkmark & \checkmark & 74 \\
E  & \textbf{\pooled{} (proposed)}         & \checkmark & \checkmark & \textbf{85} \\
\midrule
-- & \lss{} to trivial instruction text    & \checkmark & --         & 58 \\
F  & Run~E Stage~2, frozen at finetune     & \checkmark & \checkmark & 0 \\
\midrule
G  & \textbf{\dense{} (proposed)}          & \checkmark & \checkmark & \textbf{90} \\
\bottomrule
\end{tabular}
\end{table}

\subsection{Transfer and alignment granularity}
\label{sec:transfer}
The core result tests whether \lss-shaped representations generalize to tasks
\emph{not seen during Stage-2 alignment}. We run an identical Stage-3 finetune on
each held-out task from four Stage-2 backbones: R1 (H-RDT baseline), R2 (+AVP, no
\lss), R3 (+\pooled), R4 (+\dense). Because R2--R4 share the same AVP data and
differ only in the alignment objective, any R3/R4 gap over R2 is attributable to
\lss{} rather than to the human data or task matching.

\tabref{tab:transfer} shows the pattern. \dense{} (R4) gives the best result on
every task, in-distribution and held-out. \pooled{} (R3), by contrast, helps the
matched task but fails to transfer: it regresses on \shakebottle{} (39 vs.\ R2's
45) and is essentially flat on \movecanpot{} (19 vs.\ 20). This is the granularity
claim concretized---collapsing phase structure into one episode embedding
over-specializes the representation to the alignment task, whereas phase-local
alignment carries over. All baselines (R1) are reproduced on our hardware under
the identical 50-episode / 22{,}000-step protocol, so every column in
\tabref{tab:transfer} is directly comparable.

\begin{table}[t]
\centering
\caption{Transfer to held-out tasks (success rate \%, 100 rollouts). Columns are
Stage-2 backbones: R1 = \hrdt{} baseline, R2 = +human data (no \lss{}), R3 =
+\pooled{}, R4 = +\dense{}. All share an identical 50-episode / 22{,}000-step
Stage-3 finetune, so the only difference is the Stage-2 alignment objective.
\dense{} gives the best result on every task; \pooled{} helps the in-distribution
task but does not transfer.}
\label{tab:transfer}
\small
\setlength{\tabcolsep}{6pt}
\begin{tabular}{@{}lcccc@{}}
\toprule
Held-out task & R1 & R2 & R3 & R4 \\
\midrule
\adjustbottle{} (in-dist.) & 70  & 79 & 85 & \textbf{90} \\
\shakebottle{}             & 34  & 45 & 39 & \textbf{54} \\
\movecanpot{}              & 9.1 & 20 & 19 & \textbf{26} \\
\bottomrule
\end{tabular}
\end{table}

\subsection{Mechanism probe}
\begin{figure}[t]
  \centering
  \includegraphics[width=0.9\linewidth]{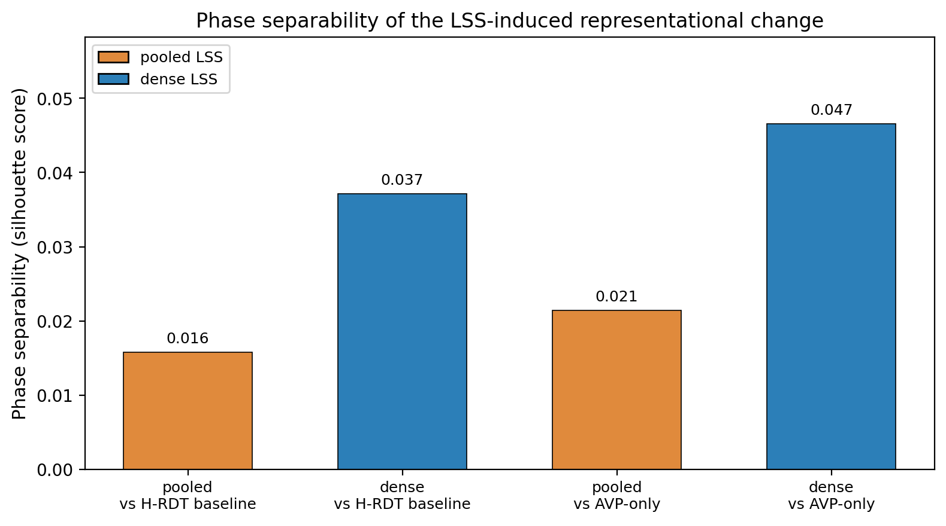}
  \caption{Per-phase separability of the \lss-induced representational change on
  \adjustbottle{}. Under the same small alignment budget ($\lambda{=}0.1$),
  \dense{} yields ${\sim}2\times$ the silhouette of \pooled{}; absolute values are
  small by design. Supporting evidence for phase-local shaping.}
  \label{fig:probe}
\end{figure}

As supporting evidence for \emph{why} \dense{} transfers, we probe the backbone's
action-token hidden states (not the discarded head) on the \adjustbottle{}
human-pretraining data and measure how strongly the \lss-induced representational
change separates by phase, using a silhouette score. \dense{} yields roughly twice
the phase separability of \pooled{} (silhouette 0.047 vs.\ 0.021 against the
AVP-only baseline; 0.037 vs.\ 0.016 against the H-RDT baseline). The absolute
values are small by design: with $\lambda{=}0.1$, \lss{} is a deliberately weak
secondary pressure that must not override action prediction, so it shifts the
representation only slightly (and the per-phase targets are themselves only
modestly distinct, mean pairwise cosine 0.70). The informative quantity is
therefore the \emph{relative} gap---under the identical small budget, dense
separates phases about twice as strongly as pooled (Fig.~\ref{fig:probe}). We
treat this as supporting,
not primary, evidence; the primary evidence is the behavioral transfer in
\secref{sec:transfer}.

\subsection{Inference cost}
A defining property of \lss{} is that the entire alignment apparatus---the \head{}
and the reasoning encoder---exists only during training and is discarded before
deployment. The deployed policy is therefore the unmodified base VLA: its
per-step latency and parameter count are identical to the \hrdt{} baseline, and it
adds \emph{zero} cost relative to a policy trained without \lss{}. This contrasts
structurally with test-time reasoning methods, which re-incur their reasoning
computation at every control step---generating reasoning tokens or predicting
future states before each action---an overhead that recurs over the horizon and
has motivated dedicated efficiency work\cite{ecot,cotvla,fastecot}. \lss{} obtains
the benefit of reasoning supervision while paying none of this recurring cost.

%% file: sections/limitations.tex
\vspace*{0.5\baselineskip}
\section{Limitations}
Several limitations scope our claims. The Stage-2 alignment uses a single human
pretraining task in a tabletop bottle-manipulation domain; broader generality
across object categories and skills is left to future work. \dense{} depends on
the quality of VLM-generated phase boundaries and rationales, for which we apply
no formal quality control; noisy segmentation can in principle misalign tokens.
Because \lss{} bakes reasoning into the weights, it offers no test-time
replanning---it trades the adaptivity of generate-at-inference methods for their
runtime cost. Finally, the mechanism probe reports small absolute separability
values and is bounded by the distinctness of the reasoning targets; we treat it
as supporting rather than decisive evidence.

%% file: sections/conclusion.tex
\vspace*{0.5\baselineskip}
\section{Conclusion}
We introduced Latent Semantic Scaffolding, a training-time auxiliary loss that
aligns a VLA's action-token representations to physical-reasoning embeddings and
is discarded at inference. Our central finding is that \emph{alignment granularity}
is the decisive lever: per-token, phase-local alignment (\dense) transfers to
held-out tasks where episode-pooled alignment over-specializes, and it does so at
zero added inference cost. A representational probe supports phase-local shaping as
the mechanism. Future work includes additional human-motion sources and object
domains, and real-robot validation.
\vspace*{0.5\baselineskip}